\documentclass[conference]{IEEEtran}
\IEEEoverridecommandlockouts

\usepackage[sort&compress, numbers]{natbib}
\usepackage{amsmath,amssymb,amsfonts}
\usepackage{algorithm}
\usepackage{algorithmic}
\usepackage{graphicx}
\usepackage{textcomp}
\usepackage{xcolor}

\def\BibTeX{{\rm B\kern-.05em{\sc i\kern-.025em b}\kern-.08em
    T\kern-.1667em\lower.7ex\hbox{E}\kern-.125emX}}

\usepackage{mathtools}
\usepackage{wrapfig}
\usepackage{lipsum}
\usepackage{bm}
\usepackage{bbm}
\usepackage{enumitem}

\newcommand{\bb}{\texttt{BBOpt}}
\newcommand{\bo}{\texttt{BO}}
\DeclareMathOperator*{\argmax}{argmax}

\newcommand{\GP}{\mathcal{GP}}
\newcommand{\K}{\mathcal{K}}

\begin{document}

\title{Collaborative and Federated Black-box Optimization: A Bayesian Optimization Perspective\\
\thanks{This work is supported by NSF CAREER Award 2144147 and the NSF CMMI-FMRG Award 2328010} }

\author{\IEEEauthorblockN{Raed Al Kontar}
\IEEEauthorblockA{\textit{Industrial \& Operations Engineering} \\
\textit{University of Michigan}\\
alkontar@umich.edu}
}

\maketitle

\begin{abstract}
We focus on collaborative and federated black-box optimization (BBOpt), where agents optimize their heterogeneous black-box functions through collaborative sequential experimentation. From a Bayesian optimization perspective, we address the fundamental challenges of distributed experimentation, heterogeneity, and privacy within BBOpt, and propose three unifying frameworks to tackle these issues: (i) a global framework where experiments are centrally coordinated, (ii) a local framework that allows agents to make decisions based on minimal shared information, and (iii) a predictive framework that enhances local surrogates through collaboration to improve decision-making. We categorize existing methods within these frameworks and highlight key open questions to unlock the full potential of federated BBOpt. Our overarching goal is to shift federated learning from its predominantly descriptive/predictive paradigm to a prescriptive one, particularly in the context of BBOpt \textemdash an inherently sequential decision-making problem.

\end{abstract}


\section{Introduction}
The tremendous increase in computational capabilities of edge devices, along with the rapid market infiltration of powerful AI chips, has led to explosive interest in collaborative and distributed analytics, such as federated learning (FL), which distributes model learning across diverse and often heterogeneous data sources to process more of the user’s data at its point of origin. FL addresses many of the privacy concerns, regulatory constraints, communication costs, and skyrocketing data volumes that have made traditional cloud-centric computation increasingly unsustainable.

Significant progress has been made in FL. Methods have been proposed to enable faster convergence \cite{karimireddy2020scaffold, nguyen2020fast}, address heterogeneity in size and distribution \cite{sattler2019robust,li2019fedmd}, improve parameter aggregation schemes \cite{wang2020federated}, personalize across concept and covariate shifts \cite{shi2023personalized,fallah2020personalized}, protect against adversarial attacks \cite{bhagoji2019analyzing}, promote fairness \cite{li2019fair,yue2023gifair}, and quantify uncertainty \cite{yue2024federated, shi2023ensemble}, among many others (see \cite{kontar2021internet} for a detailed review). To date, these efforts have focused mainly on \textit{predictive} modeling, where the goal is to create a global or personalized predictive map (often a deep network) that leverages knowledge from different sources while circumventing the need to share raw data. In addition, recent \textit{descriptive} FL literature has been proposed to better understand the shared and unique features across diverse datasets. This work focuses on distributed low-rank decomposition methods such as PCA \cite{shi2022personalized}, matrix completion \cite{shi2023heterogeneous, shi2024triple}, and dictionary learning \cite{liang2024personalized}.

Yet, a key opportunity lies in advancing FL from a primarily predictive/descriptive paradigm to a \textit{prescriptive} one, which remains in its infancy. While many avenues exist for exploration, this paper provides a vision and mathematical framework for prescriptive FL in the context of black-box optimization (\bb{}) \textemdash \textit{an inherently sequential decision making problem}. Needless to say, the success of many real-world problems critically depends on trial \& error, where the goal is to manipulate a set of variables, hereon referred to as \textit{designs}, to achieve an optimal outcome. At its core, \bb{} represents the mathematical framework for trial \& error, where the relationship between designs and outcomes is often unknown (i.e., black-box), and optimal designs can only be identified through sequential experimentation. Since experiments are often expensive, and search regions high-dimensional, the goal of \bb{} is to carefully decide on the next-to-observe design(s), in order to find a good design with the fewest trials possible.

Now, if a fleet of agents exists, federated \bb{} sets forth a collaborative approach whereby agents sequentially distribute their experimentation efforts to improve and fast-track their optimal design process. If successful, federated \bb{} can significantly reduce trial \& error cost and time and benefit all participating entities, all while circumventing the need to share raw data to preserve privacy, security, and intellectual property. Despite its appeal, fundamental challenges must be first addressed to enable federated \bb{}. 
\begin{figure}[!htbp]
\vspace{-1em}
    \centering
    \centerline{\includegraphics[width=0.82\columnwidth]{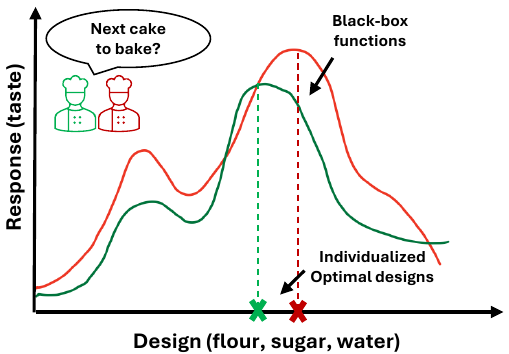}}
    \vspace{-0.5em}
    \caption{Collaborative \& federated black-box optimization}
    \label{fig:motivation} 
\end{figure}

\begin{itemize}
    \item \textbf{Challenge 1}: How to effectively distribute experimentation across collaborating entities? 
    \item \noindent \textbf{Challenge 2}: How to collaborate in the presence of heterogeneity? Heterogeneity can be both in the black-box function across agents, as well as in their resources and fidelity. Here, it is worthwhile noting that if agents are homogeneous, solutions do exist from the the rich literature on Batch \bb{} \cite{gonzalez2016batch, azimi2010batch}. 
    \item \textbf{Challenge 3}: How to design collaboration that respects the privacy of all collaborating entities? This may be critical to persuade agents to join the collaborative process.
\end{itemize}

Fig. \ref{fig:motivation} provides an anecdotal example, where two chefs collaborate to find the optimal flour, sugar and water levels for baking the best cake. While the design-response relationships share commonalities, the chefs may have slightly different palates, and they may only collaborate if their recipes and outcomes remain private. 
Given these challenges and the emerging nature of this field, this paper sheds light on mathematical frameworks that enable federated \bb{} and highlights key open questions that must be addressed to fully unlock its potential.


\section{Setting the Stage}
In the literature, \bb{} has been approached from multiple angles, including Bayesian optimization (\bo{}), derivative-free optimization, and evolutionary algorithms. While all these methods have the potential to be extended to federated settings, we primarily focus on \bo{} \cite{frazier2018tutorial}. Notably, the few existing works on federated \bb{} (which will be highlighted as we proceed) mainly fall within the realm of \bo{}. 

Mathematically, the goal of \bo{} is to find a design that optimizes a black-box function $f: \mathbb{R}^d\to\mathbb{R}$,
\begin{equation}
    \bm{x}^*=\argmax_{{\bm{x}\in\mathcal{X}\subseteq\mathbb{R}^d}} f(\bm{x}) \, ,
\end{equation}
that models the underlying true relationship between a design point $\bm{x} \in \mathbb{R}^d$ and the response $f$. Clearly, since $f$ is unknown, using first or second order optimization algorithms is not feasible, as we can only observe a potentially noisy version $y(\bm{x}) \triangleq f(\bm{x}) + \epsilon$  of $f$ by running an experiment at $\bm{x}$. These experiments, whether through physical experimentation or simulations, come with time and budget costs. As such, one needs to carefully decide on the next-to-observe design. To do so, \bo{} resorts to a utility function $U(\bm{x}):\mathbb{R}^d\to\mathbb{R}$ \cite{gramacy2009adaptive} that quantifies the benefits gained if one were to conduct an experiment at design point $\bm{x}$. 

A common example is the improvement utility function $U(\bm{x}) = \max (f(\bm{x})-y^*, 0)$ \cite{jones1998efficient} where $y^*$ is the current best observed response. Basically, $U(\bm{x})$ only gives utility to design points that give a better outcome than the current best $y^*$. Clearly, before doing an experiment, one cannot calculate the utility as we do not know $f(\bm{x})$.  Yet, at some time $t$ one can start with a small initial dataset $\mathcal{D}_t=\{(\bm{x}_1,y_1), \cdots,(\bm{x}_t,y_t)\}$ with $t$ observations. Using $\mathcal{D}_t$, a surrogate model $\mathbb{P}_{\hat{f}|\mathcal{D}_t}$ that estimates the relationship between $f$ and $\bm{x}$ can be built to predict the outcome at unobserved designs. With this, the next-to-observe design at time $t$ can be chosen as the design that gives the best utility in expectation:
\begin{align} \label{eq:original} \vspace{-0.5em}
    \bm{x}_t^{\text{new}}=\argmax_{\bm{x}} F_{t} \triangleq \mathbb{E}_{\mathbb{P}_{\hat{f}|\mathcal{D}_t}}\left[U(f(\bm{x}); \mathcal{D}_t)\right].
\end{align}
Notice that expectation is taken in (\ref{eq:original}) as the surrogate $\mathbb{P}_{\hat{f}|\mathcal{D}_t}$ is usually a posterior belief. Predominantly, \bo{} surrogates are Gaussian processes ($\GP$) or more recently Bayesian deep neural networks.

\section{Federated Frameworks}
Now, given $K$ collaborating agents, each with their own dataset $\mathcal{D}_{k,t}$ at some time $t$, we assume (for now) that each agent has a budget of $T$ experiments across $T$ iterations. Additionally, agents can communicate with each other directly or via a central orchestrator. In this collaborative setting, our goal is to enable agents to borrow strength from one another in deciding on their individualized next-to-observe designs, $\bm{x}^{\text{new}}_{k, t}$.

Next, we present three unifying frameworks that enable federated \bo{}, categorize the existing methods within these frameworks, and discuss open questions associated with each. We note that the methods introduced aim to identify the optimal $\bm{x}^{\text{new}}_{k, t}$ while preserving privacy and are repeated for all $T$ iterations until the experimentation budget is exhausted.

\subsection{Global Decisions}
The first framework is one where the the next experiments $\bm{x}^{\text{new}}_{k, t}$ are dictated by a central entity that aggregates summary statistics from all agents. Perhaps a natural start here is to follow the building-block literature in predictive FL, where the goal is to minimize the expected loss across all agents. In a federated \bo{} setting, this translates to maximizing an average over all the agents expected utility functions $F_{k,t}=\mathbb{E}_{\mathbb{P}_{\hat{f}_k|\mathcal{D}_{k,t}}}\left[U(f_k(\bm{x}); \mathcal{D}_{k,t})\right]$ at each time $t$. That is: \begin{align}
\label{eq:new} 
\max_{\bm{x}} \sum_{k=1}^Kp_k F_{k,t} =\sum_{k=1}^Kp_k\left[\mathbb{E}_{\mathbb{P}_{\hat{f}_k|\mathcal{D}_{k,t}}}\left[U(f_k(\bm{x}); \mathcal{D}_{k,t})\right]\right],  
\end{align}
where $p_k$ is some weight given to an agent $k \in [K]$, with $\sum p_k = 1$. If this was feasible, one could use the rich literature on global predictive FL, starting from the famous \texttt{FedAvg} algorithm \cite{mcmahan2017communication}, where at each communication round $r\in[R]$, agents perform few optimization iterates on $F_{k,t}$ to obtain local candidate solutions $\bm{x}^{\text{c},(r)}_{k,t} \leftarrow \texttt{agent-update}(F_{k,t})$, which are then averaged centrally to get a \textit{global} design $\bm{x}^{\text{new}, (r)}_t = \sum_k p_k \bm{x}^{\text{c}, (r)}_{k,t}$. The global design then serves as an initialization to the next optimization round, until reaching $\bm{x}^{\text{new}}_t=\bm{x}^{\text{new}, (R)}_t$. Then all agents run the experiment at $\bm{x}^{\text{new}}_t$. 


However, in contrast to predictive FL, where the global design serves as an initialization for the next round, in our prescriptive setting, $\bm{x}^{\text{new}}_t$ represents the next-to-observe design. Consequently, having all agents run the same experiment is inefficient and leads to a suboptimal use of collaborative resources. Instead, the collaborative process should allow agents to decide on their individualized tests $\bm{x}^{\text{new}}_{k, t}$.

\subsubsection{Consensus} To this end, one way to rethink federated \bo{} is through consensus, proposed in \cite{yue2023collaborative}, where each agent's next experiment $\bm{x}^{\text{new}}_{k, t}$ is a weighted combination of all agents' (including its own) candidate solutions $\{\bm{x}^{\text{c}}_{k,t}\}_{k\in[K]}$. Mathematically, this is
\begin{align}
\bm{x}_{k,t}^{\text{new}} =& \left[(\bm{W}^{(t)}\otimes\bm{I}_d)\bm{x}^{\mathcal{C}}_t\right]_k \label{eq:consensus} \\
&\text{such that} \quad 
\bm{x}^{\text{c}}_{k,t}=\left[\bm{x}^{\mathcal{C}}_t\right]_k=\argmax_{\bm{x}_k} F_{k,t} \label{eq:local} \, ,
\end{align}
where $\bm{x}^{\mathcal{C}}_{t} =  [\bm{x}^{\text{c} \top}_{1,t},  \cdots, \bm{x}^{\text{c} \top}_{K,t}]^\top$ is the concatenation of ``candidate'' solutions that maximize each agent's individual expected utility. In addition, the consensus matrix $\bm{W}^{(t)}$ is a symmetric, doubly stochastic matrix (i.e., $\sum_k w^{(t)}_{kj} = \sum_j w^{(t)}_{kj}=1$ for $j,k \in [K]$) with non-negative elements $w^{(t)}_{kj} \leq 1$, $\bm{I}_d$ is a $d\times d$ identity matrix, and $\otimes$ denotes the Kronecker product operation, i.e., $\bm{W} \otimes \bm{I}_d$ results in a $dK \times dK$ matrix.

The main idea of the formulation is intuitive. Agents first find their own utility maximizers $\bm{x}^{\text{c}}_{k,t}$, but their actual actions (experiments) $\bm{x}_{k,t}^{\text{new}}$ are a weighted combination of $\bm{x}^{\text{c}}_{k,t}$ from all agents in the system. The consensus matrix $\bm{W}^{(t)}$ dictates such weights and accordingly dictates how much one agent's decision will depend on others. Indeed, such an approach has several interesting features. (i) First, the consensus step $(\bm{W}^{(t)}\otimes\bm{I}_D)\bm{x}_{\mathcal{C}}$ naturally yields $K$ \textit{entity-specific} designs for agents to \textit{collaboratively explore and exploit} the search space. (ii) Second, the consensus step only requires each agent sharing $\bm{x}^{\text{c}}_{k,t}$; hence, information sharing is reduced to sharing designs without disclosing any agent-specific outcomes. (iii) Third, and perhaps most importantly, the consensus matrix $\bm{W}^{(t)}$ can be time-varying, allowing one to dynamically adjust the dependence of one agent's decision on another. This adds a critical layer of flexibility and enables to account for heterogeneity, which will be discussed shortly. (iv) Fourth, unlike in predictive FL (as in \texttt{FedAvg}), there are no iterations over $R$ communication rounds to find $\bm{x}_{k,t}^{\text{new}}$. Instead, agents try to find their local optimal solutions $\bm{x}^{\text{c}}_{k,t}$ once and then send them to the cloud.



These properties above yield a naturally distributed algorithm.
\begin{algorithm}[H]
\caption{Consensus iterated over $T$ Experiments}
	\label{alg::generalfl}
	\begin{algorithmic}[1]
            \STATE {\bfseries Trial:} Agents observe $\bm{y}_{k,t-1}^{\text{new}}$ at $\bm{x}_{k,t-1}^{\text{new}}$
            \STATE {\bfseries Posterior update:} Agents find $\mathbb{P}_{\hat{f}|\mathcal{D}_{t}}$ with new data
            \STATE {\bfseries Optimize:} Agents find $\bm{x}^c_{k,t}$ using (\ref{eq:local}) and send to cloud
            \STATE {\bfseries Consensus:} Cloud finds $\bm{x}_{k,t}^{\text{new}}$ using (\ref{eq:consensus}) and sends to agents
	\end{algorithmic}
\end{algorithm}
\vspace{-0.5em}

A central property of consensus is the dynamic flexibility of $\bm{W}^{(t)}$. Such a property is \textit{not only helpful but necessary} when heterogeneity exists. Specifically, when $f_1 \neq \cdots \neq f_K$, it is required for $\bm{W}^{(t)}\to\bm{I}$. To see this, consider Fig. \ref{fig:motivation} and assume the two agents have reached their optimal design. Then, any $\bm{W} \neq \bm{I}$ will always cause them to move away from their optimum. This inspires a simple framework in (\ref{eq:W1}) where the off-diagonal elements of $\bm{W}^{(t)}$ decay linearly to zero and diagonal elements to 1. A similar concept for dithering collaboration is also used in \cite{dai2021differentially,dai2020federated}, albeit within a different framework (introduced shortly). The intuition is as follows: In early stages, agent $k$ lacks enough data to build a high-quality surrogate and therefore should leverage information from others. As agent $k$ gathers more data in the later stages, they will focus more on their own objectives to find personalized optimal designs. 
\begin{align}
\label{eq:W1}
    \bm{W}^{(t+1)}=\bm{W}^{(t)}+\begin{bmatrix}\frac{K-1}{TK}&-\frac{1}{TK}&\ldots&-\frac{1}{TK}\\ \vdots & \vdots & \vdots & \vdots \\ -\frac{1}{TK}&-\frac{1}{TK}&\ldots&\frac{K-1}{TK}\end{bmatrix} \, ,
\end{align}
It should be noted that this is just one example; \textit{many other interesting ideas can be incorporated}. For instance, one could design $\bm{W}^{(t)}$ for agents to be primarily influenced by ``leaders'' that have much better designs that others (i.e., larger responses), while the diagonal elements of the leading agents are simultaneously decreased to maintain exploration. \cite{yue2023collaborative} provides such examples. 

\subsubsection{Open Questions \& Drawbacks}


The consensus framework raises many interesting questions: (i) Sharing $\bm{x}^{\text{c}}_{k,t}$ without its output may compromise privacy. While noise can be added, $\bm{x}^{\text{c}}_{k,t} + \epsilon_t$, what are the privacy-performance trade-offs? (ii) What if agents have varying fidelities? How can we incorporate this into $\bm{W}^{(t)}$ and perhaps learn the fidelities in the process? (iii) If resources vary, how can we restrict some agents' budgets to, say, $T-T'$ experiments while maintaining the effectiveness of consensus? (iv) Can we dynamically allow agents to borrow more strength from similar counterparts? Understanding heterogeneity for black-box functions is, of course, very challenging. Also, \bo{} is designed for maximization, not for fully understanding the response surface in a way that provides a measure of heterogeneity. That said, can the observed improvements across iterations guide an agent on which other agent(s) they should depend on more?

\paragraph{Note on theory} We conclude by noting that, while consensus somewhat resembles predictive FL frameworks, deriving a theoretical foundation poses a critical open question (and perhaps a drawback). First, the black-box nature of $f_k$ makes it difficult to derive theory. Second, despite recent advances in understanding the generalization error of $\mathcal{GP}$s \citep{lederer2019uniform}, understanding how these errors propagate to the, often non-concave, expected utility $F_{k,t}$ remains an open and challenging problem. 


\subsection{Conditioned Local Decisions}
Global decisions dictate the the next-to-observe designs to all agents, which may hinder collaboration in certain FL settings. Additionally, consensus requires all agents to collaborate actively and synchronously, which can be quite restrictive in many real-life scenarios. For example, consensus does not allow the use of good designs derived from historical data or human expertise.

Thus, an alternative framework for federated \bo{} is to move decision-making locally, where an agent $k$ makes local decisions conditioned on  information $E_{k,t}$ from others agents $k' \in [K \setminus k]$. Yet, what should $E_{k,t}$ be ? The hope in FL is that $E_{k,t}$ is tiny in size but rich in information.

\subsubsection{Sharing Near-optimal Designs} Following \bo{}'s philosophy to allocate designs near-optimal regions, and since we believe the black-box functions may have commonalities across agents, then they may share similar near-optimal regions. Therefore, if agents can help pinpoint each other to such regions, the problem simplifies then to finding individualized solutions within the region. With this rationale, one can set set $E_{k,t}$ be a set of \textit{synthetic} designs shared from other agents $k' \in [K \setminus k]$ to agent $k$, such that every borrowed design $\bm{x}_{k',t}^+ \in E_{k,t}$ satisfies the constraint $f_k(\bm{x}_{k',t}^+) > \delta_{k,t}$ for a constant $\delta_{n,t}$ picked by agent $k$. 

\begin{figure}[!htbp]
\vspace{-1em}
    \centering
    \centerline{\includegraphics[width=0.7\columnwidth]{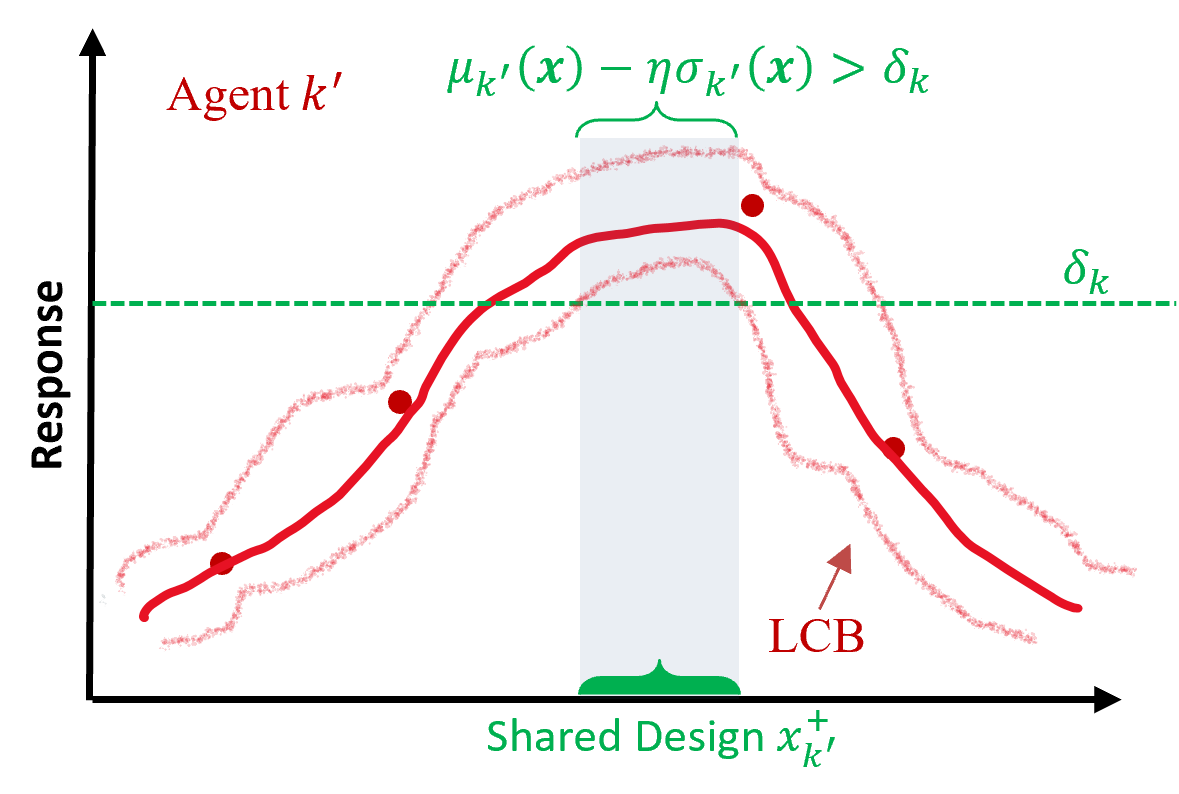}}
    \vspace{-1em}
    \caption{Agent $k'$ shares a design from the shaded region with agent $k$}
    \label{fig:constraint} 
\end{figure}

Along this line, \cite{chen2024multi}, chose $\bm{x}_{k',t}^+ \in E_{k,t}$ to be designs from other agents $k'$ that satisfy the constraint 
\begin{align}
    \label{eq: test_constr}
    \mu_{k',t}(\bm{x})-\eta_t \cdot \sigma_{k',t}(\bm{x}) > \delta_{k,t} \triangleq \max_{\bm{x}}\mu_{k,t}(\bm{x}) \, ,
\end{align}
\noindent where $\mu_{k,t}$ and $\sigma_{k,t}$ for an agent $k$ are the mean and standard deviation of $\mathbb{P}_{\hat{f}_k|\mathcal{D}_{k,t}}$. Therefore, every design $\bm{x}_{k',t}^+  \in E_{k,t}$ points out a potential design with a better response compared to agent $k$'s current best, i.e., $\max_{\bm{x}}\mu_{k,t}(\bm{x})$, according to their posterior belief. In other words, any design in $E_{k,t}$ points out to a region for potential improvement. This is shown in Fig. \ref{fig:constraint}. Critically, the comparison function in (\ref{eq: test_constr}) can be efficiently and privately calculated for FL using secure multiparty computation, which is known as Yao’s Millionaires' problem \cite{lin2005efficient}.

A simple choice that satisfies (\ref{eq: test_constr}), if it exists, is agent $k'$'s lower confidence bound (LCB) maximizer; $\bm{x}_{k',t}^+ = \argmax_{\bm{x}\in \mathcal{X}} \left\{\mu_{k',t}(\bm{x})-\eta_t \cdot \sigma_{k',t}(\bm{x})\right\}$.  It is worthwhile noting that designs (i) suggested by human experts or (ii) extracted from historical knowledge can simply be placed within $E_{k,t}$.

Now, agent $k$ presumes that for all $\bm{x}^+_{k',t} \in E_{k,t}$, there holds $f_{k}(\bm{x}^+_{k',t})>\delta_{k,t}$. Notice that this constraint may not hold because of heterogeneity or $\GP$ estimation error, but this can be readily accommodated as will become clear shortly. With $E_{k,t}$, agent $k$ will update their posterior belief by conditioning on $E_{k,t}$ to make a local decision. The aim is to \textit{fuse one's private data $\mathcal{D}_{k,t}$ with the shared knowledge $E_{k,t}$ that may be pointing to regions of improvement}. 

Mathematically, the conditioned surrogate is given as $\mathcal{F}_{k,t}^+ \triangleq \mathbb{P}_{\hat{f}_k|\mathcal{D}_{k,t}} | E_{k,t}$. However, note that $E_{k,t}$ is not a dataset but a set of constraints, which essentially leads to a constrained $\GP$. Fortunately, sampling from a constrained $\GP$ is well studied, and one can employ basic rejection sampling (RS) to do so. A key benefit of RS in a constrained $\GP$ is that one can set a finite sampling size, and if no samples are accepted at a borrowed design $\bm{x}^+_{k',t}$, the design is discarded. By doing so, \textit{agent $k$ discards borrowed knowledge that significantly contradicts its posterior belief}. This is the key to handling heterogeneity and demonstrating sublinear regret guarantees, regardless of how heterogeneous the functions are. 

By extracting samples from $\mathcal{F}_{k,t}^+$, the expected utility $\mathbb{E}_{\mathcal{F}_{k,t}^+}\left[U(f(\bm{x}); \mathcal{D}_t)\right]$ can be approximated using Monte Carlo (see \cite{chen2024multi}), and a local decision  $\bm{x}^{\text{new}}_{k, t}$ is made accordingly.

\subsubsection{Sharing Near-optimal Design Distributions}
While not currently proposed, an approach similar in spirit to sharing near-optimal designs  could involve sharing distributional beliefs over the optimal design location. Mathematically, we can let $E_{k,t} = \{\pi_{k'}\}_{k'}$, where $\pi_{k'}$ is a density shared by agent $k'$. A simple example would be $\pi_{k'} = \mathcal{N}(\bm{x}^+_{k',t}, \sigma^2 \bm{I}_d)$ for some predetermined $\sigma$. Alternatively, agent $k'$ can use repeated Thompson sampling to estimate their current belief about their optimal design, given as $\pi_{k'}(\bm{x}^*_k)$.

Now, instead of tweaking the surrogate as proposed in \cite{chen2024multi}, one can directly adjust the expected utility $F_{k,t}$ in a manner similar to incorporating prior knowledge into BO in $\pi$BO \cite{hvarfner2022pi}. With this, we can redefine the local objective to be:
\begin{align}
\label{eq:utility}
\bm{x}_{k,t}^{\text{new}} = \argmax_{\bm{x}} \left[F_{k,t} \times \left(\prod_{k'} \pi_{k'}\right)^\frac{\beta}{T}\right]
\end{align}
The main idea here is to tweak the local objective $F_{k,t}$ using beliefs about the optimal design in $\pi_{k'}$. Given potential heterogeneity, the impact of information shared by other agents decays at a rate of $\beta/T$ for some constant $\beta$. This allows an agent to give greater weight to designs that others consider optimal, while eventually prioritizing its own objective. Various adjustments to this approach are possible, such as taking a weighted product of the ${\pi_{k'}}$ values, with each $\pi_{k'}$ decaying at a different rate. Additionally, privacy can be preserved in a federated setting by adding additive or multiplicative noise to the shared density or its summary statistics.

With this approach, it is evident that updating the utility directly, rather than altering the surrogate, is mathematically simpler. However, it introduces the challenge of tuning $\beta$, which should be a function of the similarities across $f_k$ and $f_{k'}$'s. Given the black-box nature of the functions, this can be a pathological problem. In contrast, the proposed approach of tweaking the surrogate $\mathbb{P}_{\hat{f}_k|\mathcal{D}_{k,t}} | E_{k,t}$ does not depend on tuning parameters and comes with strong guarantees simply since it enables discarding shared information that significantly conflict with an agent $k$'s own posterior belief.

\subsubsection{Sharing $\GP$ Features} Within the general framework of sharing and utlizing lightweight information $E_{k,t}$ for local decisions, \cite{dai2020federated} tackled federated \bo{} by sharing $\GP$ Random Fourier Features (RFFs) \cite{rahimi2007random}. Using RFFs, stationary kernels ($\K(\bm{x}, \bm{x}') = \K (\bm{x}-\bm{x}')$) can be approximated by a set of random features $\bm{\phi}(\bm{x}) \in \mathbb{R}^D$; where $\K(\bm{x}, \bm{x}')=\bm{\phi}^\top(\bm{x})\bm{\phi}(\bm{x})$. Therefore, the black-box function is approximated using Bayesian linear regression as $\hat{f}_k(\bm{x})=\bm{\phi}^\top(\bm{x}) \bm{w}_k$ such that  the posterior of $\bm{w}_k$ assuming a prior $\bm{w}_k \sim \mathcal{N}(0, \bm{I}_D)$ is  
\begin{align}
    \mathbb{P}_{\bm{w}_k|\mathcal{D}_t} = \mathcal{N}(\bm{\nu}_{k,t}, \sigma^2 \bm{\Sigma}_{k,t}^{-1}) \, , \notag
\end{align}
where \(\bm{\Phi}_{k,t} = [\bm{\phi}(\bm{x}_{k,1}), \dots, \bm{\phi}(\bm{x}_{k,t})]^\top\) is a \(t \times D\)-dimensional matrix, $\bm{\Sigma}_{k,t} = \bm{\Phi}_{k,t}^\top \bm{\Phi}_{k,t} + \sigma^2 \bm{I}$, $\bm{\nu}_{k,t} = \bm{\Sigma}_t^{-1} \bm{\Phi}^\top \bm{y}_{k,[t]}$. Here $\sigma^2$ is the variance of $\epsilon$ and $\bm{y}_{k,[t]} =[y_{1,t}, \cdots, y_{k,t}]^\top$.

With this, \cite{dai2020federated} proposed that $E_{k,t}=\{\bm{w}^{(s)}_{k'}; k'
\in [K \setminus k] \}$, where $\bm{w}^{(s)}_{k'} \sim \mathbb{P}_{\bm{w}|\mathcal{D}_{k',t}}$. This entails sharing the $\mathcal{GP}$ coefficients, assuming $\bm{\phi}$ is fixed across all agents. Now, agent $k$ will either use their own $\bm{w}^s_k$ with probability $p_t$ to find $\bm{x}^{\text{new}}_{k,t}$ using the Thompson sampling utility, i.e., $\bm{x}^{\text{new}}_{k,t} = \arg\max_{\bm{x}} \bm{\phi}^\top(\bm{x}) \bm{w}^{(s)}_k$, or use a $\bm{w}^{(s)}_{k'}$ randomly sampled from $E_{k,t}$ with probability $1 - p_t$ to get $\bm{x}^{\text{new}}_{k,t}$. Similar to the consensus idea, $p_t \rightarrow 1$ so that agents eventually focus on their own objectives. This idea was further expanded in \cite{dai2021differentially} to guarantee differential privacy in FL by sharing an averaged coefficient $E_{k,t}=\{\bar{\bm{w}}_{k' \subset [K \setminus k]}\}$, which is clipped to bound its $L_2$ norm and has Gaussian noise added to it. The local process remained the same.

\subsubsection{Open Questions \& Drawbacks} While sharing optimal designs can incorporate any utility and provide theoretical guarantees despite the heterogeneity level, RS scales poorly with shared designs. Furthermore, sharing $\GP$ RRFs is restricted to Thompson sampling and requires fixed features across all agents. Fortunately, the conditioned local design framework, where decisions are made at the agent level and informed by $E_{k,t}$, is \textit{generic}, allowing for various ideas to be incorporated. Some open questions include: (i) When adding noise to $E_{k,t}$ (for RFFs and designs), what are the accuracy-privacy trade-offs? (ii) Can we include upper confidence bounds information when sharing designs to gauge which shared designs have the best potential? (iii) How can non-myopic BO literature be used to enhance collaboration? (iv) Can we theoretically identify informative conditions for when collaboration is helpful to develop algorithms that determine when to stop collaborating?

\subsection{A Predictive Solution: Improving the Surrogate}
The third framework does not tackle decision-making directly but instead aims to improve the local predictive model (i.e., the \bo{} surrogate $\mathbb{P}_{\hat{f}_k|\mathcal{D}_{k,t}}$) through collaboration. Local decisions then follow the regular \bo{} process as in (\ref{eq:original}). A central feature of this framework is that it can potentially be incorporated within the first two. For instance, consensus can be achieved using surrogates learned in a federated and collaborative manner. Here, two ideas arise: 

\subsubsection{Federated $\mathcal{GP}$} In $\bo{}$, $f_k$ is often modeled as a $\GP$ where, $f_k(\bm{x})\sim \GP (0,\K(\cdot,\cdot;\bm{\theta}_{\K}))$ and $y_k=f_k(\bm{x})+\epsilon_k$ where $ \epsilon \overset{\text{i.i.d.}}{\sim} \mathcal{N}(0, \sigma^2)$. $\bm{\theta} = (\bm{\theta}_{\K},\sigma^2)$ parameterize the $\GP$ and encode our prior belief. Naturally,  $\bm{\theta}$ could be learned in a federated fashion. Here the utility function $L_{k,t}$ for each agent is the log-marginal likelihood (\ref{eq:gp}). Now, in FL, the goal is to collaboratively learn $\bm{\theta}$ that maximizes the global utility \vspace{-0.5em}
\begin{align} \label{eq:gp}
    \argmax_{\bm{\theta}}\sum_{k=1}^Kp_k L_{k,t}& \triangleq \notag \\ &\sum_{k=1}^Kp_k \log \mathbb{P} (\bm{y}_{k, [t]}|\mathcal{D}_{k,t};\bm{\theta}) \, .
\end{align}

Recently, \cite{yue2024federated} proposed \texttt{FedAvg} to solve (\ref{eq:gp}). Unlike in deep learning, $L_{k,t}$ is a log-likelihood function featuring \textit{correlations} (due to the $\GP$), which leads to stochastic gradients being biased estimators of the full gradient. Nevertheless, \cite{yue2024federated} shows that despite these correlations, \texttt{FedAvg} converges to a critical point of (\ref{eq:gp}), subject to statistical errors. A key advantage of a federated $\mathcal{GP}$ is its inherent ability to personalize, as estimating $\bm{\theta}$ is equivalent to learning a global $\GP$ prior. Personalized predictions $\mathbb{P} (y^*_k|\mathcal{D}_{k,t}, \bm{x}^*;\hat{\bm{\theta}})$ are obtained by conditioning on a local agent's data $\mathcal{D}_{k,t}$. The local data inherently fulfill the role of personalization.

\subsubsection{Federated Multi-output $\mathcal{GP}$ ($\mathcal{MGP}$)} Alternatively, one may give more model flexibility by assuming agents have shared $\bm{\theta}^g$ and unique parameters $\bm{\theta}^k$ through an $\mathcal{MGP}$ where \vspace{-1em}
\begin{align}
    [f_1 (\bm{x}), \cdots, f_K (\bm{x})]^\top \sim \mathcal{MGP} (\textbf{0}, \bm{\text{cov}}^f_{k,k'}(\bm{x},\bm{x}')) \, .
\end{align}
 Many options can be used to construct $\mbox{cov}^f_{k,k'}(\bm{x},\bm{x}')$. Perhaps, the most common approach is convolution processes \cite{alvarez2011computationally}
$f_k(\bm{x})=\sum_{q=1}^{Q}\int_{-\infty}^{\infty}\K_{q,k}(\bm{x} - \bm{u}; \bm{\theta}^k) g_q(\bm{u}; \bm{\theta}^g) d \bm{u}$ where $g$ is a latent $\mathcal{GP}$. The key advantages is that multiple latent functions $g_q$ allow information sharing across different agents through different kernels $\K_{q,k}(\bm{x})$.

Unlike federated $\GP$s, the global $\mathcal{MGP}$ utility cannot be written as a sum of agent utilities like (\ref{eq:gp}), since agents are all correlated by construction. Despite this, \cite{chung2024federated} recently show that, using variational inference, one can approximate the joint $\mathcal{MGP}$ log-likelihood as $\sum_{k=1}^K p_k \tilde{L}_{k,t}$, where $\tilde{L}_{k,t}$'s are independent. With this, they developed a simple $\texttt{FedAvg}$-like approach to update $\{\{\bm{\theta}^k\}, \sigma^2, \bm{\theta}^g\}$ simultaneously, while only sharing $\bm{\theta}^g$ with the orchestrator.

\subsubsection{Key Open Question} A central question for integratively learning surrogates is: What is the best way to initialize federated \bo{}? Should agents begin with space-filling designs, or should they explore different space partitions? Furthermore, how will this affect $\GP$ estimation error early on?

\section{Conclusion}
We present a vision and unifying frameworks for collaborative and federated black-box optimization from a Bayesian optimization perspective while highlighting the many exciting problems yet to be addressed. We conclude by noting that many applications could benefit from this emerging literature, including collaborative optimization of manufacturing process parameters, material discovery, hyperparameter tuning and clinical trials. Therefore, we believe a high-impact endeavor would be to bring this theory to practical applications by providing a working example where agents collaborate to improve their decisions and benefit from collaboration, all while preserving their privacy and intellectual property.

\bibliographystyle{IEEEtran}

{\footnotesize
\bibliography{Raed_cited}}

\end{document}